\documentclass[preprint,12pt]{elsarticle}

\usepackage{amsmath,amsfonts,bm}

\def\eqref#1{equation~\ref{#1}}

\def\floor#1{\lfloor #1 \rfloor}
\def\1{\bm{1}}

\DeclareMathAlphabet{\mathsfit}{\encodingdefault}{\sfdefault}{m}{sl}
\SetMathAlphabet{\mathsfit}{bold}{\encodingdefault}{\sfdefault}{bx}{n}

\newcommand{\be}{\begin{eqnarray}}
\newcommand{\ee}{\end{eqnarray}}
\newcommand{\bee}{\begin{eqnarray*}}
\newcommand{\eee}{\end{eqnarray*}}

\newcommand{\matrixb}{\left[ \begin{array}}
\newcommand{\matrixe}{\end{array} \right]}

\usepackage{lineno,hyperref}
\usepackage{adjustbox}
\usepackage{url}
\usepackage[utf8]{inputenc} 
\usepackage[T1]{fontenc}    
\usepackage{hyperref}       
\usepackage{url}            
\usepackage{booktabs}       
\usepackage{amsfonts}       
\usepackage{nicefrac}       
\usepackage{microtype}      
\usepackage{amsmath}
\usepackage{amsthm}
\usepackage{amssymb}
\usepackage{graphicx}
\usepackage{xspace}
\usepackage{tabularx}
\usepackage{multicol}
\usepackage{multirow}
\modulolinenumbers[5]

\journal{Neurocomputing}

\begin{document}

\begin{frontmatter}

\title{Understanding Dropout as an Optimization Trick} 

\author{Sangchul Hahn}
\author{Heeyoul Choi \fnref{fn1}}
\fntext[fn1]{Corresponding Author: heeyoul@gmail.com}
\address{Dept. of Information and Communication Engineering, \\
Handong Global University, South Korea 37554}

\begin{abstract}
As one of standard approaches to train deep neural networks, dropout has been applied to regularize large models to avoid overfitting, and the improvement in performance by dropout has been explained as avoiding co-adaptation between nodes. However, when correlations between nodes are compared after training the networks with or without dropout, one question arises if co-adaptation avoidance explains the dropout effect completely. In this paper, we propose an additional explanation of why dropout works and propose a new technique to design better activation functions. First, we show that dropout can be explained as an optimization technique to push the input towards the saturation area of nonlinear activation function by accelerating gradient information flowing even in the saturation area in backpropagation. Based on this explanation, we propose a new technique for activation functions, {\em gradient acceleration in activation function (GAAF)}, that accelerates gradients to flow even in the saturation area. Then, input to the activation function can climb onto the saturation area which makes the network more robust because the model converges on a flat region.  Experiment results support our explanation of dropout and confirm that the proposed GAAF technique improves image classification performance with expected properties.
\end{abstract}


\end{frontmatter}


\section{Introduction}

Deep learning has achieved state-of-the-art performance or eventually surpassed human-level performance on many machine learning tasks, such as image classification, object recognition, and machine translation \citep{He2015resnet, Redmon2016yolo9000, Vaswani2017}. To achieve such impressive performance, many techniques have been proposed in different areas: optimization (e.g., Adam \citep{Kingma2014adam} or Adadelta \citep{zeiler2012adadelta}), regularization (e.g., dropout \citep{Hinton2012dropout}), activation function (e.g., $ReLU$ \citep{Nair2010}), or layer (e.g., batch-normalization \citep{Ioffe2015}, Resnet \citep{He2015resnet}). Even with great success, many techniques and algorithms are not yet well-understood, and different hypotheses are proposed to explain how they work or improve the performance \cite{Zhang2017, Baldi2013, Karpathy2015}. 

To train deep neural networks, regularization is crucial to prevent the models from overfitting and to increase generalization effect for new data samples. As a regularization method, dropout was proposed to prevent co-adaptation among the hidden nodes of deep feed-forward neural networks by dropping out randomly selected hidden nodes \citep{Hinton2012dropout, srivastava2014dropout}. Co-adaptation is a process by which two or more nodes behave as if they are a single node or group, which emerges by interaction as response to the backpropagation in training process. When some nodes are updated and behave together with increased interdependence, the model might lose a part of its computational power and the partial gradient in backpropagation becomes more inaccurate compared to the total gradient. Dropout is known to break the ties by dropping some of them randomly. When dropout is analyzed with probabilistic models, dropout is assumed to avoid the co-adaptation problem \citep{Baldi2013,Kingma2015,Gal2016}. However, we believe that the co-adaptation avoidance is not the only reason to explain why dropout improves performance significantly in many applications. 

On the other hand, it is known that when the outputs of the nonlinear activation function are saturated, the loss function can converge onto a flat region rather than a sharp one with a higher probability \citep{Hochreiter1997flat, Chaudhari2016}. Flat regions provide better generalization for test data, because some variation in the input data cannot create a significant difference in the output of layers. However, it is usually hard to train neural networks with input in the saturation areas of the nonlinear activation functions because there is no gradient information flowing in the areas. That is, there is a {\em dilemma of nonlinearity} to train neural networks since saturation areas are necessary to make the forward propagation effective while such areas put the backward propagation in trouble with zero gradient.

In this paper, we raise a question on the conventional explanation of dropout, if the effect of dropout is completely explained by avoiding co-adaptation. What else would partly explain why dropout works? Furthermore, if a new additional explanation is persuasive, there is a chance for developing another learning techniques based on such explanations.  
Basically, our hypothesis is that dropout can be partly explained as an efficient optimization technique, so to handle the dilemma between forward and backward propagation which shall be described in the next section. 
We show that dropout makes more gradient information flow even in the saturation areas and it pushes the input towards the saturation area of the activation functions by which models can become more robust after converging on a flat region.   

Based on the additional explanation, we propose a new technique for the activation function, {\em gradient acceleration in activation function (GAAF)} that directly adds gradients even in the saturation areas, while it does not change the output values of a given activation function. Thus, GAAF makes models to obtain a better generalization effect. In GAAF, gradients are explicitly added to the areas where dropout generates gradients, so that GAAF makes gradients flow through layers in a deterministic way, contrary to dropout which makes gradients stochastically. Thus, GAAF trains neural networks with less iterations than dropout does. 

The paper is organized as follows. Background knowledge including dropout is described in Section 2. In Section 3, we provide an additional explanation about how dropout works in terms of optimization. We propose a new technique, GAAF, for activation functions in Section 4. The experiment results are presented and analyzed in Section 5, followed by Section 6 where we conclude.

\section{Background}
In this section, we briefly review nonlinear activation functions, dropout, and noise injection in neural networks. 

\subsection{Nonlinear Activation Functions in Neural Networks}
In fully connected neural networks, one layer can be defined as follows: 
\be
h_j & = & \phi (z_j),\\
z_j & = & \sum_i W_{ij} x_i + b_j,
\label{eq:activation}
\ee
where $\phi(\cdot)$ is a nonlinear activation function such as $sigmoid$, $tanh$, and $ReLU$. $x_i$ and $h_j$ are input and output for the layer, and $W_{ij}$ and $b_j$ are weight and bias, respectively. The sum $z_j$ is referred to as a {\em net} for the output node $j$. That is, $z_j$ is an input to $\phi(\cdot)$. 

In backpropagation, to obtain error information for the node $j$ in the current hidden layer $H_l$, the derivative of $\phi(z_j)$ is multiplied to the weighted sum of the errors $\delta_k$ from the upper layer $H_{l+1}$ as defined in Equation (\ref{eq:bprop}). 
\be
\delta_j & = & \phi'(z_j) \sum_{k \in H_{l+1}} W_{jk} \delta_k. 
\label{eq:bprop}
\ee

Note that $\delta_j$ approaches zero when $\phi'(z_j)$ is close to zero, and the amount of gradient information for the weights connected to the node $j$ is proportional to $\delta_j$. In other words, when a net value is close to the saturation areas of $\phi(\cdot)$, gradient information vanishes and the connected weights cannot be updated. While the saturation areas hinder training, the functions can play an important role (i.e., nonlinear transformation) around the saturation areas where $\phi(z_j)$ actually provides the nonlinear property. This is the dilemma of nonlinearity in neural networks. That is, it is important for the net values to go into saturation areas in forward propagation but it is hard to train the networks to have the net values in the saturation areas in backpropagation. 

\subsection{Dropout}
Since dropout was proposed in \cite{Hinton2012dropout} to prevent co-adaptation among the hidden nodes of deep feed-forward neural networks, it has been successfully applied to many deep learning models \citep{Dahl2013icassp,srivastava2014dropout}. This method randomly omits (or drops out) hidden nodes with probability $p$ (usually $p=0.5$) during each iteration of the training process, and only the weights that are connected to the surviving nodes are updated by backpropagation. The forward propagation with dropout is defined as follows: 
\be
z_j & = & \sum_i W_{ij} d_i x_i + b_j, 
\label{eq:drop}
\ee
where $d_i$ is drawn independently from the Bernoulli distribution with probability $p$. When $d_i$ is zero, the input node $x_i$ is dropped out.  

After a model with $N$ hidden nodes is trained with dropout, to test new samples, the nodes of the model are rescaled by multiplying ($1- p$) to all the nodes, which has the effect of taking the geometric mean of $2^N$ dropped-out models. 
In \cite{Hinton2012dropout,srivastava2014dropout,Sainath2013asru}, it is shown that the neural networks trained with dropout have excellent generalization capabilities and achieve the state-of-the-art performance in several benchmark datasets \citep{Schmidhuber2015nn}. In addition to feed-forward layers, dropout can be applied to the convolutional or the recurrent layers. To preserve the spatial or temporal structure while dropping out random nodes, spatial dropout \citep{Tompson2015cvpr} and RnnDrop \citep{Moon2015} were proposed for the convolutional and the recurrent layers, respectively. 
There are several papers that explain how dropout improves the performance \citep{Baldi2013,Kingma2015,Gal2016}, assuming that dropout avoids the co-adaptation problem without any question on it. 

Interestingly, it was pointed out that batch-normalization could eliminate the need for dropout for performance and they both work towards the same goal as regularizers \cite{Ioffe2015}. If batch-normalization could eliminate the need for dropout, then they are partially playing the same role and one might argue that batch-normalization might reduce co-adaptation by the randomness of the samples in the batch. 

In this paper, we believe that the conventional explanation on the dropout effect may not be complete and that dropout might be partially explained as an effective optimization technique. 



\subsection{Noise Injection to the Network}
Like the L1 or L2 norms, regularizers can prevent models from overfitting and improve generalization capability. It has been known that adding noise during training is equivalent to regularizing the model \citep{Bishop1995nc}. In addition to dropout, there are several methods to train neural networks with noise, including weight noise injection \citep{Graves2013interspeech}, denosing auto-encoder \citep{Vincent2008}, and dropconnect \citep{Wan2013}. Those methods add (or multiply) Gaussian (or Bernoulli) noise to weight (or node) values. For example, weight noise injection adds Gaussian noise to weight values, and dropout multiplies random values drawn from the Bernoulli distribution to node values. Such methods improve performance in many tasks \citep{Graves2013interspeech, Pham2014}. 

On the other hand, noise can be applied to activation function as in noisy activation function \citep{Gulcehre2016noisy}. Noisy activation function adds noise where the node output would saturate, so that some gradient information can be propagated even when the outputs are saturated. Although noisy activation function trains the network with noise, it is not explicitly considered as a regularizer. We understand dropout in the same line with noisy activation function, that is, dropout makes gradient flow even in the saturation areas, which is described in the next section.

\section{Dropout for Optimization}
In this section, we argue that dropout can work as an effective optimization technique by making more gradient information flow through nonlinear activation functions. 

The amount of gradient information can be measured by the average of the absolute amount of gradient in each layer. We calculated the gradient information at $k$-th layer, $G_k$, with the following equation. 
\be
G_k=\frac{1}{N}\sum_{n=1}^{N}(\frac{1}{I*J}\sum_{i=1}^{I}\sum_{j=1}^{J}\left | \frac{\partial E_n}{\partial W_{i,j}^k} \right |),
\label{eq:grad_info}
\ee
where $N$ is the number of nodes in the layer, $E_n$ is the cost function, and $W_{i,j}^k$ is the weight matrix of the $k$-th layer. To confirm our argument, we compare how much gradient information flows during training process with or without dropout. Table \ref{table:grad} summarizes the amount of gradient information with the MNIST dataset during training feed-forward neural networks (512-256-256-10) with or without dropout. We can see that dropout increases the amount of gradient information around five times. 

\begin{table}[ht]
\caption{The amount of gradient information flowing through layers during training with MNIST on feed-forward networks. The values in the table are the average of the absolute value of gradient of all nodes in each layer during the whole iterations.}
\label{table:grad}
  \centering
  \begin{tabular}{|l|c|c|}
\hline
 & Without Dropout & With Dropout \\
\hline
Layer3	 & 9.35e-05  & 5.83e-04 \\
Layer2	 & 1.40e-04  & 6.52e-04 \\
Layer1	 & 1.07e-04  & 5.93e-04 \\
\hline
\end{tabular}
\end{table}
    
Then, the next question would be how dropout increases the amount of gradient information. We take a clue from how noisy activation function works \citep{Gulcehre2016noisy}, where the noise allows gradients to flow easily even when the net is in the saturation areas. We believe that dropout could increase the amount of gradient in a similar way.
We explain how dropout can increase gradient flow in two steps. First, dropout introduces variance to the net values, then the variance generates gradient flow even on the saturated areas so that the total gradient flow increases.

In deterministic neural networks, the net values are determined with zero variance. However, dropout makes a significant variance for the net, due to the randomness of $d_i$. Given the dropout probability $p$ with fixed $W_{ij}$ and $x_i$ for forward propagation, the variance can be calculated by  
\be
Var(z_j) &=& Var(\sum_i^N W_{ij} d_i x_i+b_j) \nonumber \\
  &=& p (1-p) \sum_i^N (W_{ij}x_i)^2  \gg 0. 
\ee

Variance by dropout can be empirically confirmed. The node variances from the model for MNIST trained with dropout are summarized in Table \ref{table:var}. To check the variance of net value, $z_j$, we obtained the net values for the same batch 20 times with different random dropout masks during training when the model almost converged. Then, we calculated the variance for the net value of each node and took the average of the variances in each layer. Table \ref{table:var} presents the average of net variances for one batch (128 data samples) in each layer. Note that the variance of `Layer1' (input layer) is zero, since there is no dropout in the input layer, and `Last Layer' has a variance generated by dropout in `Layer3'. 

\begin{table}[ht]
\caption{The average of net variances in each layer during training with dropout.}
\label{table:var}
  \centering
  \begin{tabular}{|l|c | c |}
\hline
 & Net Variance \\
\hline
Last Layer & 1.97 \\
Layer3	 & 1.07 \\
Layer2	 & 1.07 \\
\hline
\end{tabular}
\end{table}

\begin{figure}[ht]
\centerline{\hbox{ \includegraphics[width=2.75in]{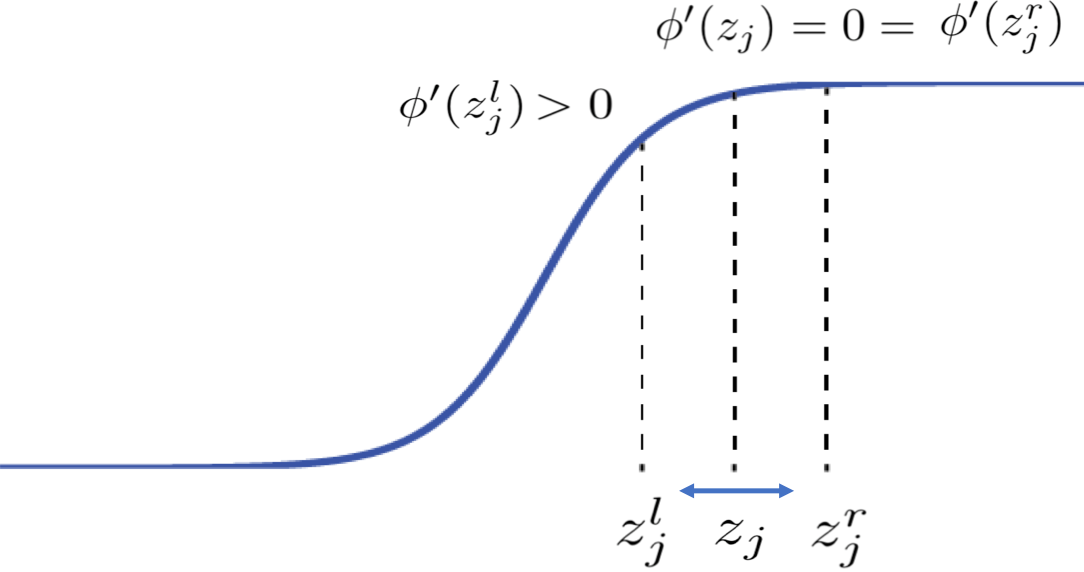}}}
\caption{Variance in the saturation area moves the net value to the left or right side, which increases the probability to have more gradient information.}
\label{fig:dropout_variance}
\end{figure}

As the second step, we describe how the variance can help increase the amount of gradient information. The variance of the net values increases chances to have more gradient information especially around the boundary of saturation areas. In Figure \ref{fig:dropout_variance}, when the derivative $\phi'(z_j)$ is (almost) zero without dropout, there is no gradient flowing through the node. However, if it has a variance, $z_j$ can randomly move to the right or left. In Figure \ref{fig:dropout_variance}, when $z_j$ moves to the right, there is no gradient information as before, but when it moves to the left, it obtains gradient information which is generated by dropout. That is, with a certain amount of probability, gradient information can flow even for $z_j$ in the figure. We believe that this phenomenon can explain the dropout effect.

\begin{figure}[ht]
\centerline{\hbox{ 
\includegraphics[width=1.6in,height=1.9in]{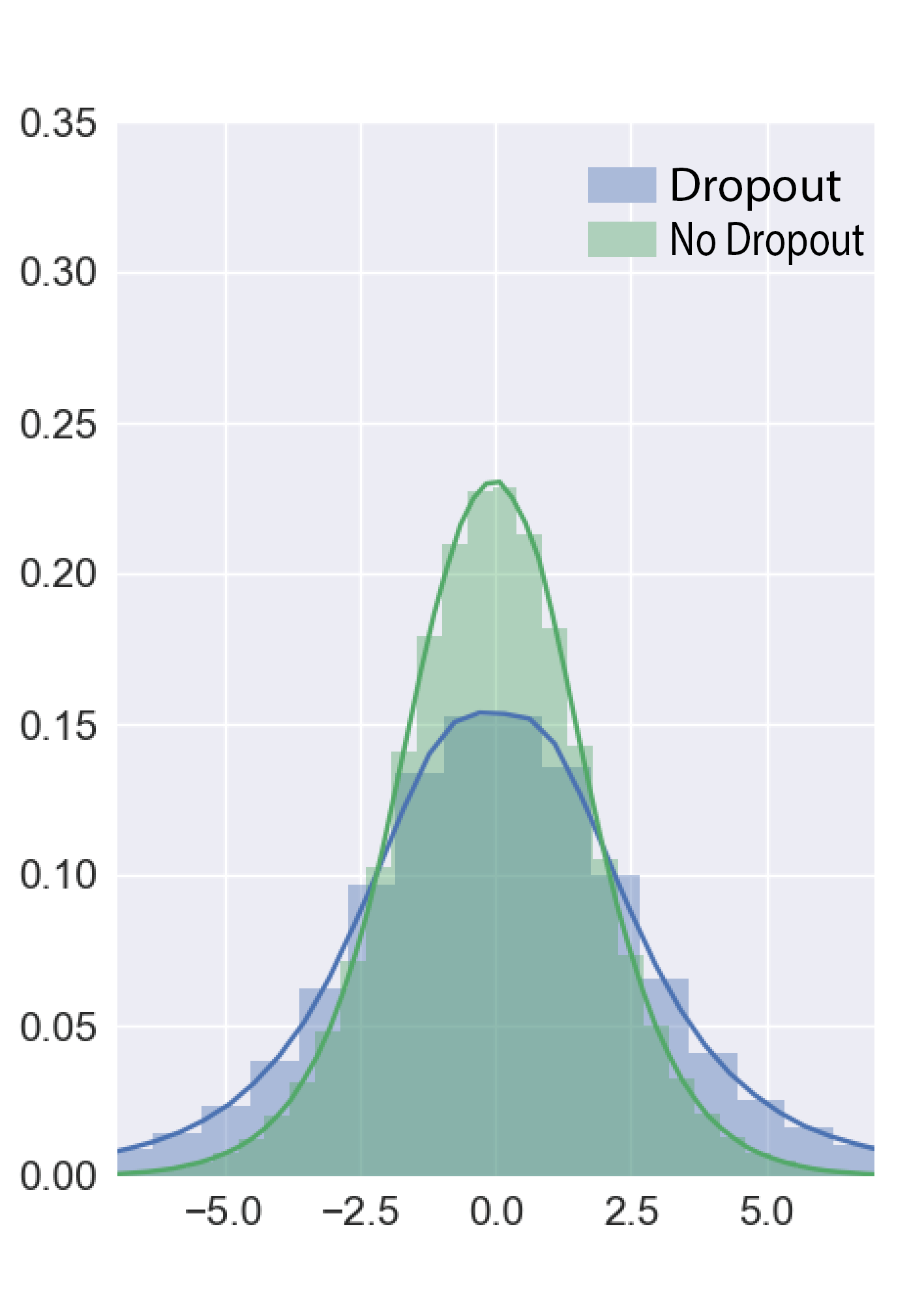}
\includegraphics[width=1.6in,height=1.9in]{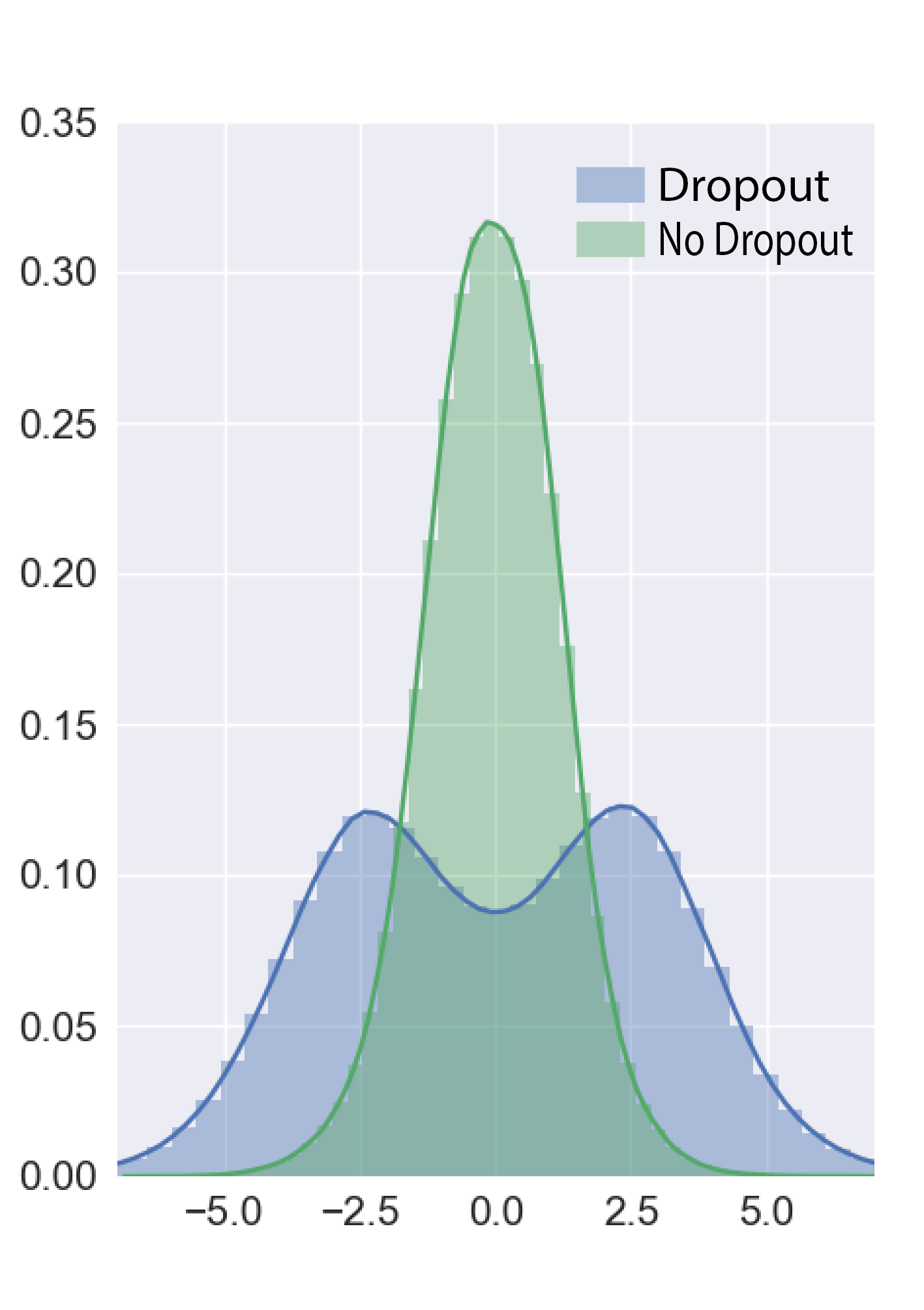}
\includegraphics[width=1.6in,height=1.9in]{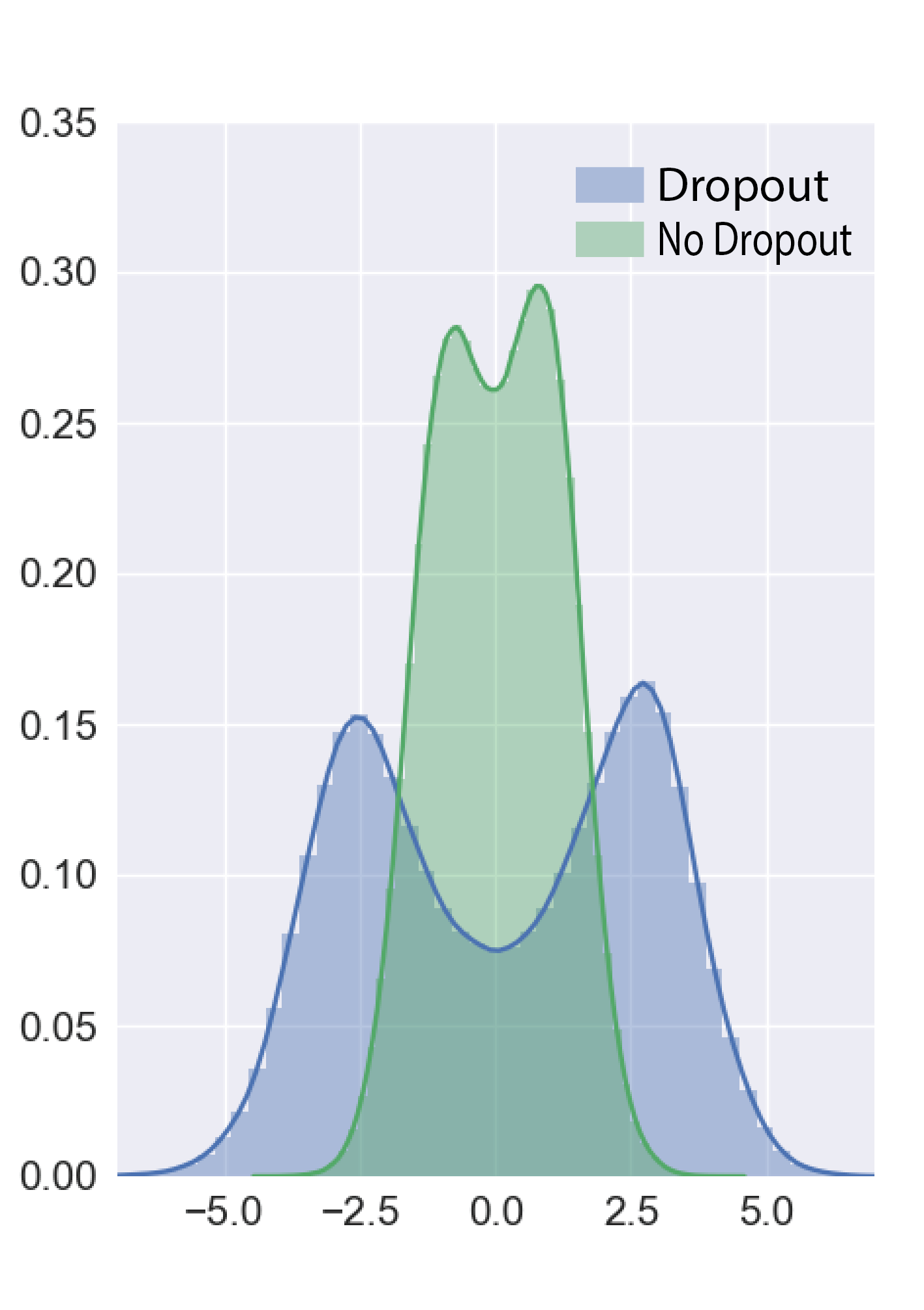}
}}
\centerline{\hbox{ 
(a) Layer 1 \hspace{0.9in} (b) Layer 2 \hspace{0.9in} (c) Layer 3}}
\caption{Distributions of net values to $tanh$ for the MNIST test data. Note that dropout pushes the net values more towards the saturation areas.}
\label{fig:node_dist}
\end{figure}

To see whether dropout actually pushes the net values towards the saturation areas, we checked the node value distributions with test data after training. Figure \ref{fig:node_dist} presents the difference between distributions of net values for MNIST test data after training with and without dropout. The model trained with dropout has more net values in the saturation area of $tanh$, which is critical to have better generalization for test data. Interestingly, the higher layer has more net values in the saturation area, since the variance of the lower layers can be transferred to the higher layer. 

\section{Gradient Acceleration in Activation Functions} 
The new understanding of dropout suggest a new trick to increase the amount of gradient information flowing through layers instead of dropout. In other words, we want a new trick to reduce training time while achieving dropout effect, since dropout takes a lot of time to train the whole networks. The idea is to directly add gradient information for the backpropagation, while (almost) not changing the output values for the forward-propagation. We call the new technique, {\em gradient acceleration in activation function} (GAAF). 

\begin{figure}[h!]
\centerline{\hbox{ 
\includegraphics[width=3.5in]{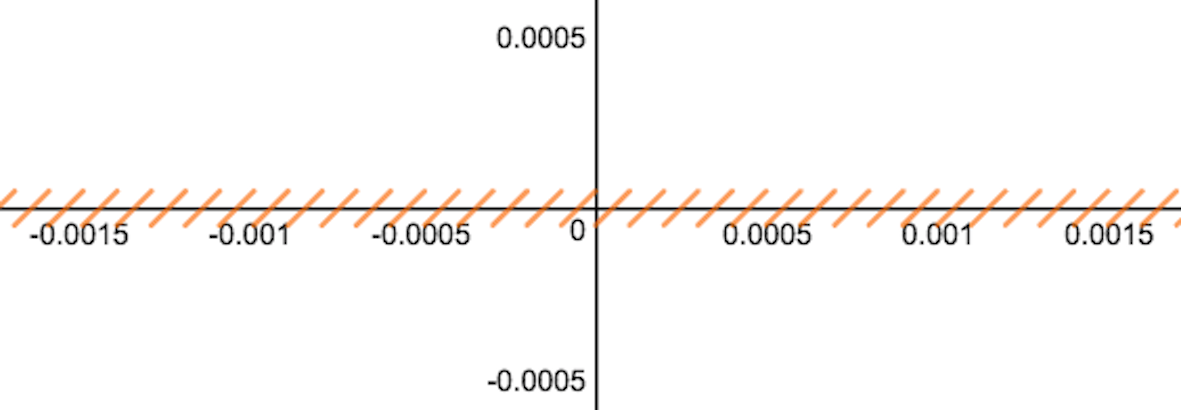}
}}
\caption{Gradient acceleration function, $g(x)$, which is drawn by the slash lines.}
\label{fig:gaf}
\end{figure}

Given a nonlinear activation function, $\phi(\cdot)$, we modify it by adding a {\em gradient acceleration function}, $g(\cdot)$ which is defined by 
\be
g(x) &=& (x*K - \floor{x*K} -0.5)/K,
\label{eq:ga}
\ee
where $\floor{\cdot}$ is the floor operation, and $K$ is a frequency constant (10,000 in our experiments). Note that the value of $g(x)$ is almost zero ($<\frac{1}{K}$) but the gradient of $g(x)$ is 1 almost everywhere, regardless of the input value $x$. The difference between $\phi(x)$ and the new function $\phi(x) + g(x)$ is less than $\frac{1}{K}$, which is negligible. Figure \ref{fig:gaf} presents what $g(x)$ looks like. 

\begin{figure}[h!]
\centerline{\hbox{ 
\includegraphics[height=0.9in]{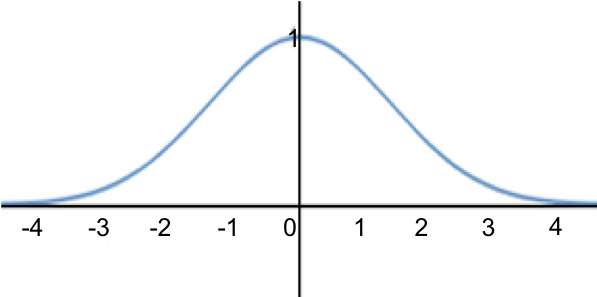}
\hspace{0.5in}
\includegraphics[height=0.9in]{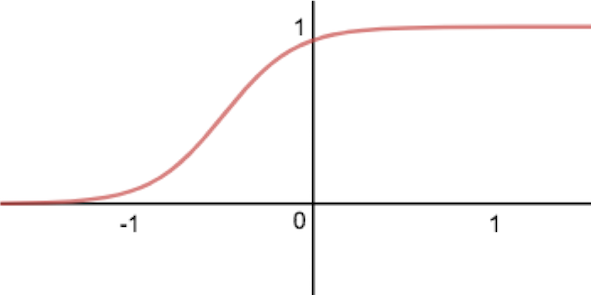}
}}
\centerline{\hbox{ 
(a) A shape function for $tanh$ \hspace{0.5in} (b) A shape function for $ReLU$}}
\caption{Shape functions for (a) $tanh$ and (b) $ReLU$.}
\label{fig:shape_func}
\end{figure}

As dropout does not generate gradient information on the leftmost or rightmost saturation areas, we also decrease the gradient acceleration on those areas by multiplying a shape function $s(\cdot)$ as in Fig. \ref{fig:shape_func} to $g(\cdot)$, which leads to our new activation function as follows: 
\be
\phi_{new}(x) &=& \phi(x) + g(x)*s(x),
\label{eq:gaaf2}
\ee
where $s(\cdot)$ needs to be defined properly depending on the activation function, $\phi(\cdot)$. For example, when $\phi$ is $tanh$ or $ReLU$, an exponential function or a shifted sigmoid function can work well as $s(\cdot)$, respectively, as shown in Figure \ref{fig:shape_func}. Basically, GAAF can be applied to all kinds of activation functions with a little adjustment of the shape function, which depends on where the saturation areas are located in the activation function. 

The proposed gradient acceleration function $g(\cdot)$ generates gradients in a deterministic way, while dropout generates gradient stochastically based on the net variances. Thus, GAAF has the same effect as dropout but it converges faster than dropout. In line with that interpretation, we can understand the different dropout rates \cite{Ba2013}. Generally, if the rate of dropout decreases, then the net variance would decrease, which in turn decreases the amount of gradient on the saturation areas. To obtain the same effect with GAAF, the shape function $s(\cdot)$ needs to be reshaped according to the dropout rate. 

\section{Experiments}
We evaluate GAAF on several image classification datasets: MNIST \citep{mnist_lecun}, CIFAR-10, CIFAR-100 \citep{cifar-dataset}, and SVHN \citep{svhn-dataset}. The MNIST dataset has hand written digit images (60K train images and 10K test images) of 10 classes (0-9 digits). The CIFAR-10 and CIFAR-100 datasets have 50K train images and 10K test images of 10 and 100 classes, respectively. The SVHN dataset has color images of house numbers from Google Street View and consists of 73K train images and 26K test images of 10 classes (0-9 digits). We use some simple models like DNN for the MNIST and VGG16 \citep{Simonyan14_vgg} for the CIFAR and SVHN dataset. In these experiments, our goal is to check if our GAAF can improve the performance of models, instead of achieving the state-of-the-art results.

\subsection{MNIST}
To evaluate GAAF on MNIST, we compared three different models: base model, dropout model, and GAAF model. The models have the same architecture, consisting of four feed-forward layers (512-256-256-10) with the $tanh$ activation function. GAAF uses an exponential function as shape function for $tanh$. Table \ref{table:mnist_result} summarizes test accuracies and the number of training epochs for each model to converge. The proposed GAAF model improves the test accuracy as much as the dropout model, while it needs less training epochs than the dropout model does.

\begin{table}[ht]
\caption{Experiment results on MNIST. The accuracies and epochs are the average values of five executions. The numbers in the parentheses are the corresponding standard deviations.}
\label{table:mnist_result}
  \centering
  \begin{tabular}{|l|c|c|c|}
\hline
 Model & Activation & Test Accuracy (\%) & Train Epochs \\
\hline
Base Model	 & $tanh$ & 98.23 (0.075) & 82 (16.6) \\
+Dropout & $tanh$ & 98.40 (0.034) & 169 (9.7)	   \\
+GAAF & $tanh$ & 98.35 (0.059) & 114 (24.8) \\
\hline
\end{tabular}
\end{table}

\begin{figure}[ht]
\centerline{\hbox{ 
\includegraphics[width=1.6in,height=1.9in]{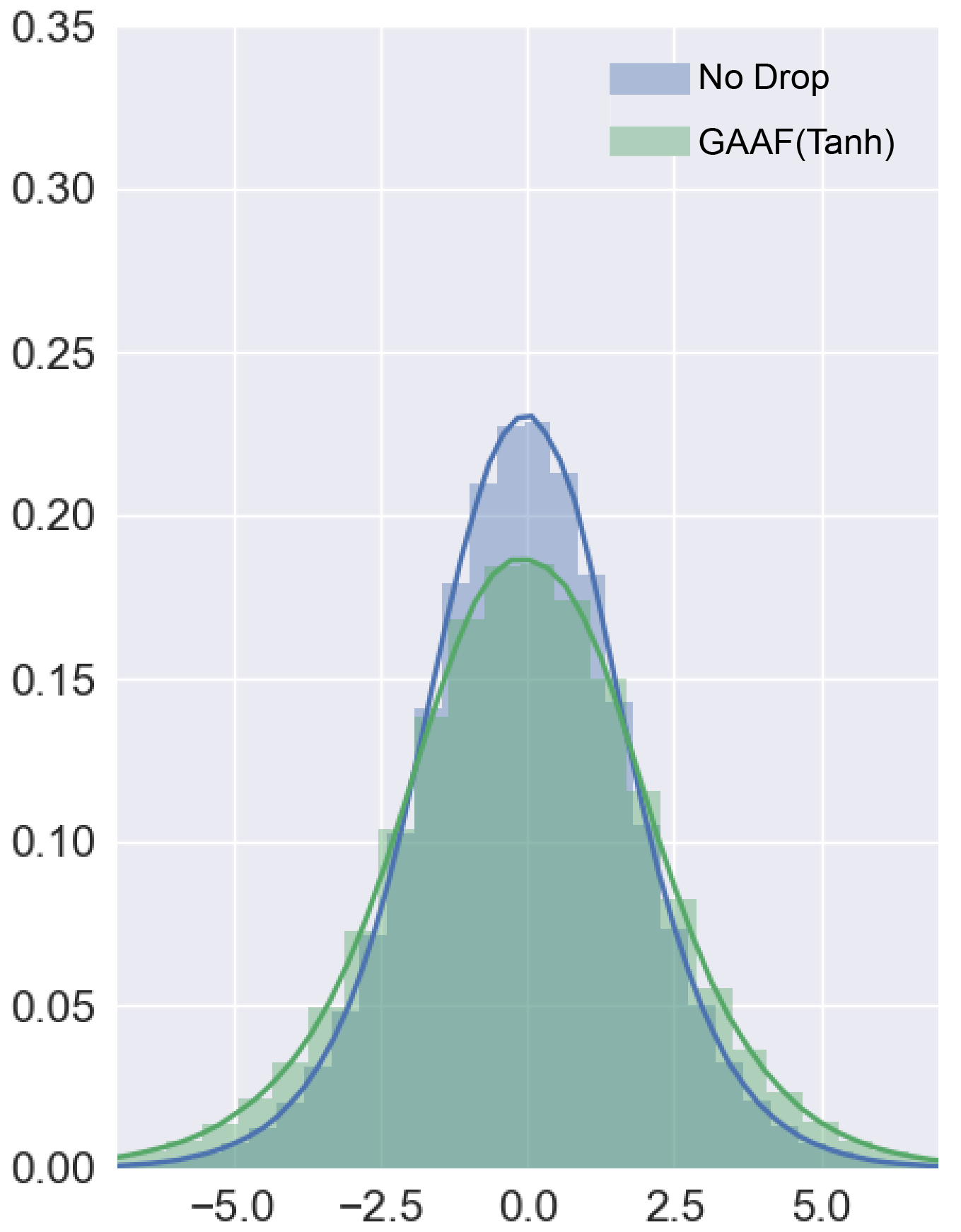}
\includegraphics[width=1.6in,height=1.9in]{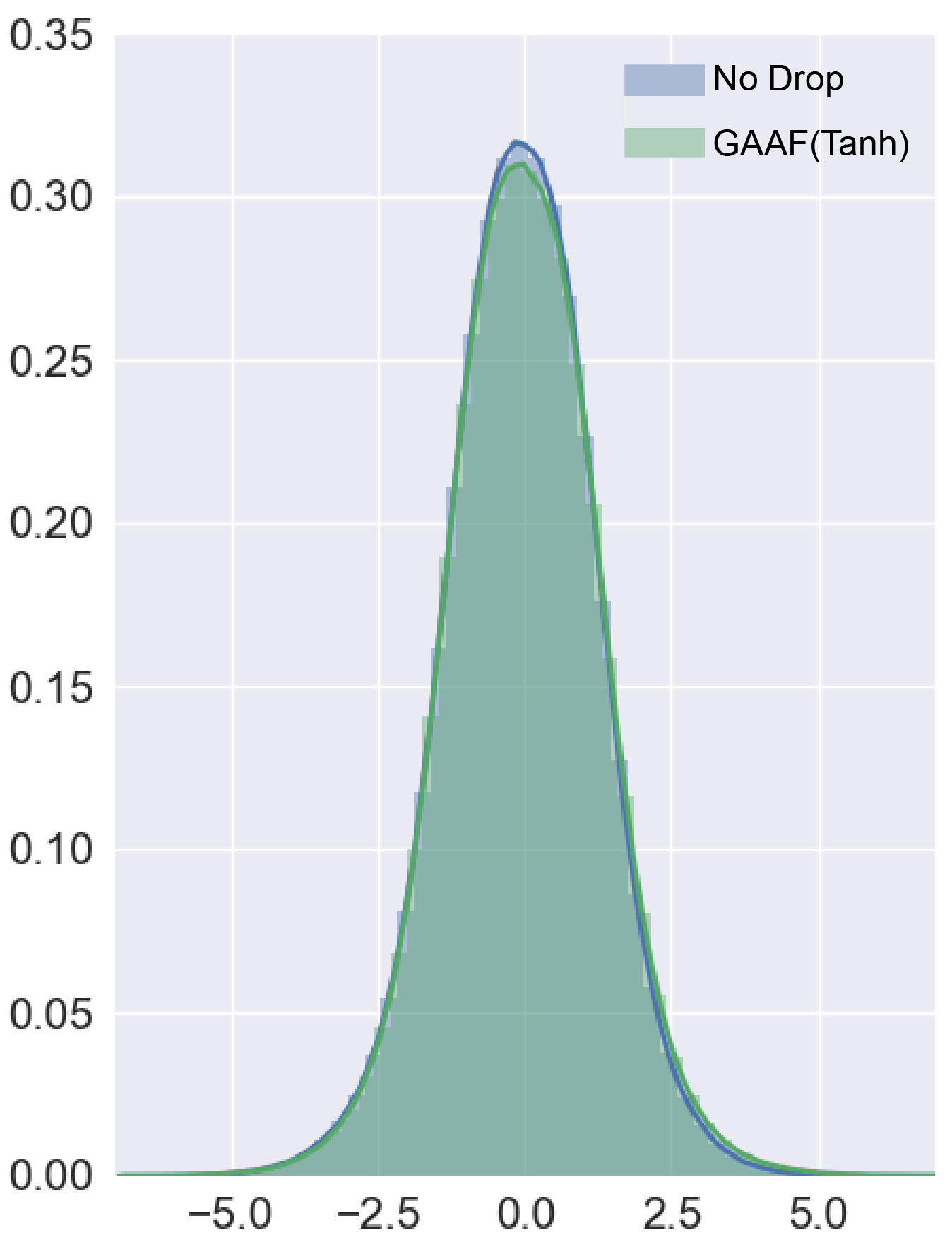}
\includegraphics[width=1.6in,height=1.9in]{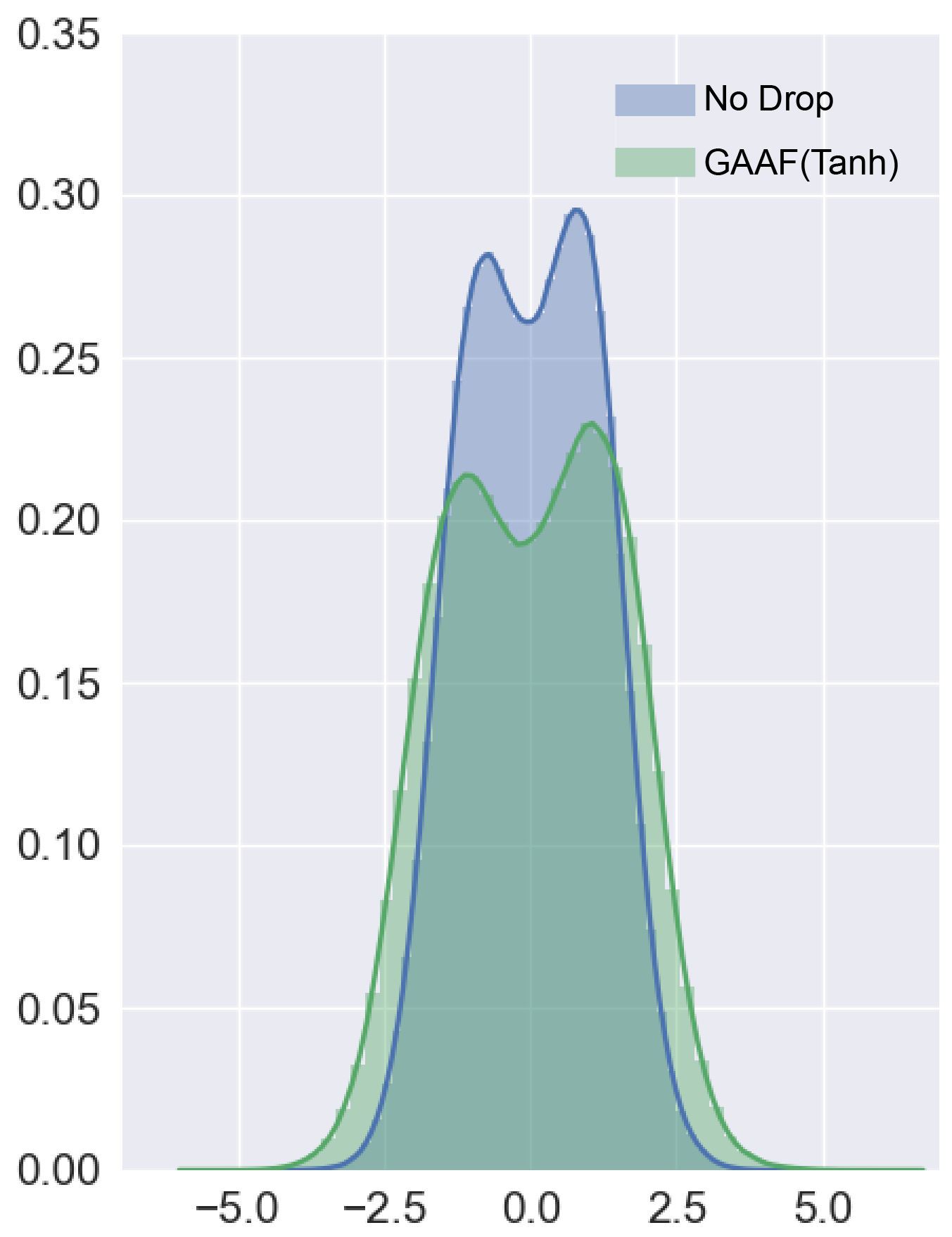}
}}
\centerline{\hbox{ 
(a) Layer 1 \hspace{0.9in} (b) Layer 2 \hspace{0.9in} (c) Layer 3}}
\caption{Distributions of net values to $tanh$ for the MNIST test data.}
\label{fig:node_gaaf}
\end{figure}

In addition, as expected, GAAF increased the gradients flowing through the layers as follows: 1.15E-3, 6.43E-4, 1.78E-4 for the three layers. Compared to the numbers in Table \ref{table:grad}, the amounts of gradients by GAAF are as large as by dropout, which are much greater than the amounts by the base model without dropout. Also, we compared the GAAF model and the model without dropout with the the distribution of net values. Figure \ref{fig:node_gaaf} shows that GAAF pushes the net to the saturation area. Although the difference is not as large as in dropout, but this seems enough to achieve the same level of performance as dropout. Also, by changing the shape function, we expect that we can push the net further towards saturation areas, which is our future work.  

\subsection{CIFAR and SVHN}
For the CIFAR datasets, we designed a base model similar to VGG16 model \citep{Simonyan14_vgg}. The model architecture is the same as the original VGG16 model, except that we removed the last three CNN layers and reduced the number of hidden nodes in the feed-forward layers, because the CIFAR image size is much smaller than ImageNet image size. 
For the SVHN dataset, we used a simple CNN model as a base model. It has four CNN layers with max pooling after every second CNN layer, and three feed-forward layers on top of the CNN layers.

We evaluated four different models: base model, base model with batch normalization \citep{Ioffe2015}, GAAF model, and GAAF model with batch normalization. Table \ref{table:cifar_svhn_result} summarizes the experiment results on the CIFAR and SVHN datasets. We used $ReLU$ as the activation function for the CNN and feed-forward layers. Thus, GAAF uses a shifted sigmoid function as shape function for $ReLU$. 

\begin{table}[h!]
\caption{Test accuracy (\%) on CIFAR and SVHN. The numbers are Top-1 accuracies. The improvements achieved by GAAF are presented in the parentheses.}
\label{table:cifar_svhn_result}
\begin{adjustbox}{width=1\textwidth}
\small
\centering
\begin{tabular}{|l|c|c|c|c|}
\hline
Model & Activation & CIFAR100 & CIFAR10 & SVHN  \\
\hline
Base Model & $ReLU$ & 59.63 & 89.55 & 92.03 \\
+Batch Norm (BN) & $ReLU$ & 67.48 & 91.1 & 93.80 \\
+GAAF & $ReLU$ & 61.29(+1.66) & 90.16(+0.61) & 92.19(+0.16) \\
+BN +GAAF & $ReLU$ & \textbf{69.36(+1.88)} & \textbf{91.92(+0.82)} & \textbf{94.16(+0.36)} \\
\hline
\end{tabular}
\end{adjustbox}
\end{table}

The results confirm that our proposed GAAF improves the base model's performance. More interestingly, GAAF improves performance even with batch normalization, contrary to dropout whose need is eliminated by batch normalization. This shows that GAAF works independently of batch normalization (maybe other optimization techniques too), while dropout hinders batch normalization (or other optimization techniques) by dropping out some (usually the half) of nodes. 

In addition, after training, the base and GAAF models have almost the same training accuracies (98.2\%, 99.6\%, and 99.9\% for CIFAR100, CIFAR10 and SVHN, respectively), while GAAF has better test accuracies as shown in Table \ref{table:cifar_svhn_result}. This supports that GAAF converges on a flat region by pushing the nets towards the saturation areas to achieve better generalization.

\section{Conclusion}
Dropout has been known to regularize large models to avoid overfitting, which was explained by avoiding co-adaptation. In this paper, we presented an additional explanation that dropout works as an effective optimization technique to generate more gradient information flowing through the layers so that it pushes the nets towards the saturation areas of nonlinear activation functions. This explanation enriches our understanding on how neural networks work. 

Based on this explanation, we proposed {\em gradient acceleration in activation function (GAAF)} that accelerates gradient information in a deterministic way, so that it has a similar effect to the dropout method, but with less iterations. In addition, GAAF works well with batch normalization, while dropout does not. Experiment analysis supports our explanation and experiment results confirm that the proposed technique GAAF improves performances. GAAF can be applied to other nonlinear activation functions with a correspondingly redesigned shape function.

\section*{Acknowledgement}
This research was supported by Basic Science Research Program through the National Research Foundation of Korea(NRF) funded by the Ministry of Education (2017R1D1A1B03033341), and by Institute for Information \& communications Technology Promotion(IITP) grant funded by the Korea government(MSIT) (No. 2018-0-00749, Development of virtual network management technology based on artificial intelligence).

\section*{References}
\bibliographystyle{elsarticle-num}
\bibliography{hchoi2019}

\end{document}